\documentclass[pmlr]{jmlr}

\usepackage{longtable}
\usepackage{tcolorbox}
\usepackage{bbm}
\usepackage{booktabs}
\usepackage[load-configurations=version-1]{siunitx} 
\usepackage{amsmath}
\usepackage{color, colortbl}
\usepackage{todonotes}

 \definecolor{Gray}{gray}{0.9}

\theorembodyfont{\upshape}
\theoremheaderfont{\scshape}
\theorempostheader{:}
\theoremsep{\newline}

\jmlrvolume{85}
\jmlryear{2019}
\jmlrworkshop{}

\title[Nonlinear Semi-Parametric Models for Survival Analysis]{Nonlinear Semi-Parametric Models for Survival Analysis}

\author{\Name{{Chirag Nagpal}, {Rohan Sangave}, {Amit Chahar}, Parth Shah,\\ Artur Dubrawski  \& Bhiksha Raj } \\
     \Email{chiragn@cs.cmu.edu,\\ \{rsangave, achahar, pvshah\}@andrew.cmu.edu,\\ \{awd, bhiksha\}@cs.cmu.edu}\\
      \addr{ Carnegie Mellon University, Pittsburgh, PA, USA}
     }

\begin{document}

\maketitle
\vspace{-3em}
\begin{tcolorbox}
 \centering {\textsf{TL;DR: stacking Non-Linearities is not always a good idea!}  }
\end{tcolorbox}

\begin{abstract}
Semi-parametric survival analysis methods like the Cox Proportional Hazards (CPH) regression \citep{cox1972regression} are a popular approach for survival analysis. These methods involve fitting of the log-proportional hazard as a function of the covariates and are convenient as they do not require estimation of the baseline hazard rate. Recent approaches have involved learning non-linear representations of the input covariates and demonstrate improved performance. In this paper we argue against such deep parameterizations for survival analysis and experimentally demonstrate that more interpretable semi-parametric models inspired from mixtures of experts perform equally well or in some cases better than such overly parameterized deep models. 
\end{abstract}

\section{Introduction}
Survival analysis aims at estimating the time of event of interest for a subject given the covariates of the subject. Cox-regression is a popular method for estimating survival functions by fitting the log-hazard rates as a linear parameterization of the individual covariates. Cox regression eliminates the need of estimating the baseline hazard and involves maximizing a partial likelihood function that is independent of the baseline hazard function. 

\noindent However, the most basic implementation of Cox-regression involves log-linear parameterization.  This is somewhat restrictive as a single linear function may not be enough to describe an entire dataset. In the light of this, recent research \cite{katzman2018deepsurv} have demonstrated that incorporating non-linearities through the use of multi-layer perceptrons helps improve performance of such survival analysis models.

\noindent Another natural approach to mitigate the restrictive effect of linearity is by using a mixture of experts \cite{jacobs1991adaptive,jordan1994hierarchical} where a softmax function determines which `expert' to use for each datapoint. (The softmax is conditioned on the individual observation).  \cite{rosen1999mixtures} propose a more appropriate version of the mixture of experts for Cox-Regression models and describe a procedure to perform parameter inference. In this paper, we argue against their proposed inference procedure and instead propose an approach motivated from modern mean field variational methods which we attempt to justify both theoretically and via extensive experimentation.


\noindent We further compare the performance of the model and our proposed inference procedure to modern non-parametric methods like Random Survival Forests \citep{ishwaran2008random} and deep parametric neural network approaches \citep{katzman2018deepsurv} and demonstrate its effectiveness on multiple real world datasets.

\begin{tcolorbox}
We hope that these contributions would spark debate in the survival analysis community and amongst healthcare practitioners of whether we really need deep neural networks for survival analysis?
\end{tcolorbox}


\paragraph{Technical Significance}
The use of deep learning for survival analysis has received attention in the community recently. One successful implementation was `Deep-Surv' \citep{katzman2018deepsurv}. In this paper we argue that non-linearities in survival data can be incorporated in a more transparent and interpretable manner by the use of the mixture of linear experts model, which was first proposed for survival data by \cite{rosen1999mixtures}. We modify their proposed optimization objective in order to make parameter learning tractable and we justify this modification both theoretically and experimentally. We further compare performance of our approach to state of the art on multiple real world survival datasets.


\paragraph{Clinical Relevance}
Interpretability in the context of machine learning systems may be defined as the ability to explain a model's decisions to humans in understandable terms such that the generated interpretation maximizes a user’s intended utility \citep{doshi2017towards, dhurandhar2017tip}. In this sense we lose out on interpretability when rationalizing the decisions of deep non-linearly parameterized models to domain experts to support decision making, in our case, clinicians. Our methodology preserves interpretability owing to the fact that the 'experts' that make predictions are linear functions of the covariates. Although non-linearities may be introduced in the selection process of our experts, we argue that our methodology is still more transparent than highly parameterized models as the weights of each covariate that are used in predictions are visible, therefore allowing for easier decision making on the part of users of the model.

\section{Preliminaries}
Survival data typically consists of triplets of the form of $( x_i, e_i,  t_i ) \in \mathcal{D}$ where $x_i$ are the covariates (features) associated with an individual $i$, $t_i$ refers to the time the event took place, and $e_i$ is a binary variable that indicates if the data is right censored, that is if the event did actually occur or if that is the last observation available for the individual. Survival analysis thus involves estimation of two quantities.\\

\noindent The \textbf{Survival Function} $S(t)$ and the baseline hazard $\lambda(t)$. The survival function $S(t)$ detemrines the probabilty of an individual surviving beyond a certain time $t$, that is $S(t)=P(T>t)$.\\

\noindent The \textbf{Hazard Function} $\lambda(t)$ on the other hand represents the individuals risk of death at a given time $t$. Thus the hazard maybe be defined as 

\begin{align}
    \lambda(t) = \lim \limits_{\delta \to 0}{}\frac{P(t\leq T < t+\delta |T \geq t)}{\delta} 
\end{align}

\noindent Cox-like regression models assume the baseline hazard $\lambda_0(t)$, or the aggregate risk shared by every individual at the time $t$ to be constant across all individuals and do not condition the baseline hazard on the individual covariates of an individual drawn from $\mathcal{D}$. That is, the hazard of an individual at a time $t$, given the covariates is modelled as
$$\lambda(t|x_i) = \lambda_0(t) \exp( \beta^\top x_i )$$
Where $\beta$ are parameters to perform inference over.

\noindent  The \textbf{Risk Set}, $\mathfrak{R}(t_i)$ or the set of all individuals who are available (have survived) at time $t_i$ maybe defined as $R(t_i) := \{ (x_j, t_j, e_j) \in \mathcal{D} | t_j \geq t_i  \}$ \\

\noindent Under the simplest Cox Proportional Hazards assumptions,  the probability that an event occurs for a subject $x_i$ at time $t_i$ conditioned on the Risk Set, $\mathfrak{R}(t_i)$ as 
\begin{align}
\text{Pr}(e_i| x_i, \mathfrak{R}(t_i), \beta ) = \frac{\exp(\beta^{\top}x_i)}{\sum \limits_{\mathfrak{R}(x_i)}{} \exp(\beta^{\top}x_j) }
\label{eq:condprob}
\end{align}
\noindent We write the quantity in Eq. \ref{eq:condprob} as a conditional as originally proposed by Cox.\footnote{Such a treatment is controversial. Subsequent literature uses the more appropriate term `Partial Likelihood'. Nevertheless, in order to make our modelling assumptions consistent with the proposed approaches, we will consider it a conditional.} 

Note that we assume no ties, that is no two subjects share the same $t_i$ and hence in the rest of the paper, we rewrite the Risk Set as a function of $x_i$ instead of $t_i$.

\section{Incorporating Non-Linear Feature Interactions}

Notice that the set model parameters, $\beta$ interact with the input features of each subject $x$, through a linear function. Deep Surv \citep{katzman2018deepsurv} instead choose to replace the interaction of $\beta$ and $x$ through the use of Multi-Layer Perceptrons, and demonstrate an improved performance in terms of Concordance-Index. A similar direction was explored by \cite{sargent2001comparison, mariani1997prognostic, xiang2000comparison} although their experiments did not demonstrate much benefit of using non-linearities through neural networks.
\begin{figure}[!h]
    \centering
    \includegraphics[width=0.5\textwidth, trim={6cm 9cm 8cm 0.4cm},clip]{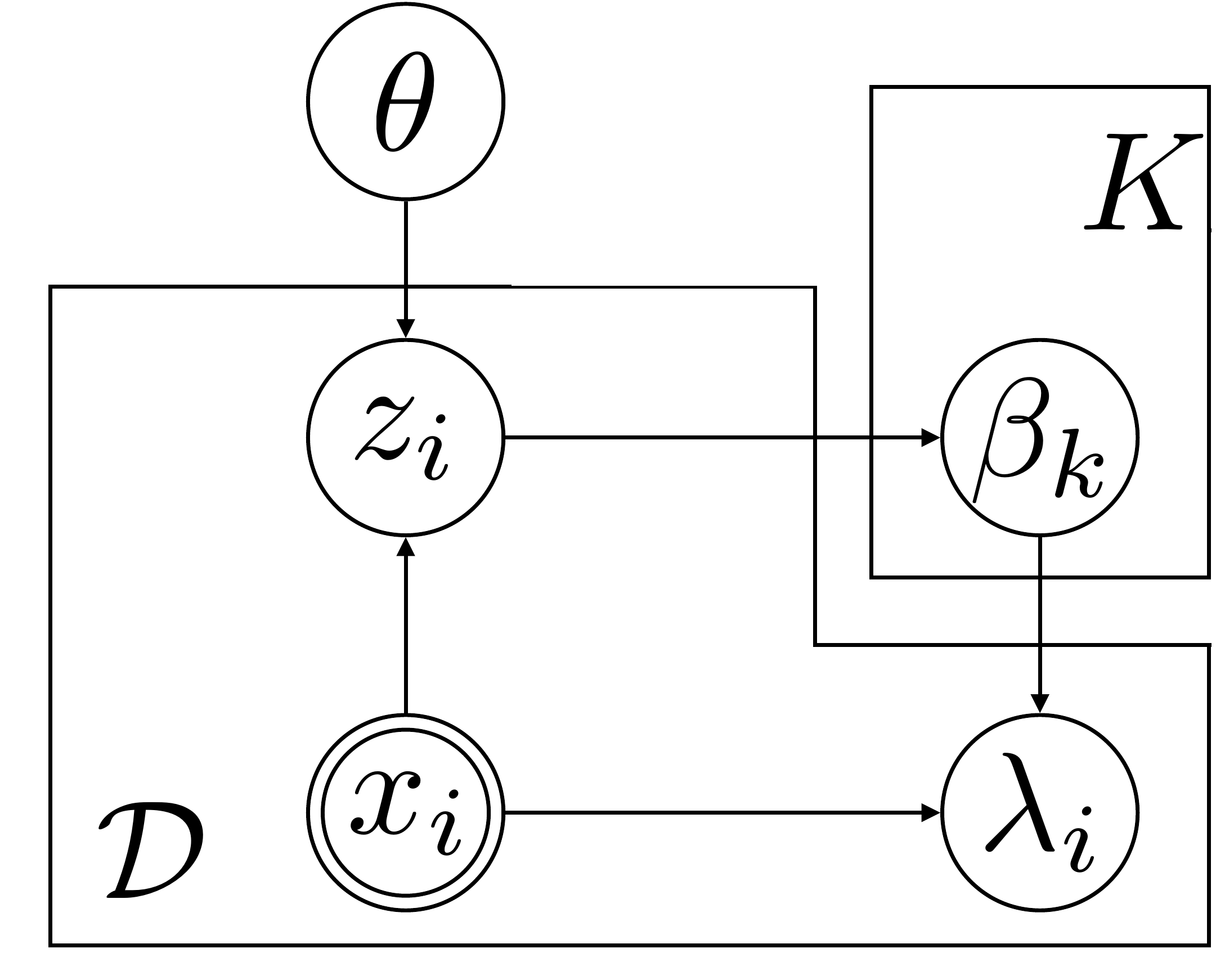}
    \caption{The Cox Mixture of Experts Model in plate notation. $x_i$ are the observed covariates for an individual $i$. $\theta$ interacts with the covariates to determine $z_i$, the latent variable that selects one of the $K$ experts. The final hazard $\lambda_i$ is an outcome of $x_i$ and the selected expert $k$ with the parameters, $\beta_k$.}
    \label{fig:pgm}
\end{figure}

\noindent In this paper we instead focus on the Mixture of Experts  (\cite{jacobs1991adaptive}), that consists of a gating function that separates the feature space into regions where different parameter sets are involved in the final outcome. The Mixture of Experts was modified for Cox regression in~\cite{rosen1999mixtures}. Their approach involves a softmax gating function that decides which set of experts to operate on each subject and a linear expert that determines the final risk corresponding to each individual as described in Fig. \ref{fig:pgm}.

 From the Directed Acyclic Graph assumptions in Fig.~\ref{fig:pgm}, it is clear that the the partial likelihood, as in Eq. \ref{eq:likelihood}, requires us to marginalize over all the latent variables $z_j$ over each element in the risk set. Such a computation is of the order of $K^{|\mathfrak{R}(x_i)|}$. Where $|\mathfrak{R}(x_i)|$ is the number of elements in the Risk Set of $x_i$ and $K$ is the number of experts. Clearly, such a computation is intractable and requires approximate techniques to perform inference. Thus under this model
\begin{align}
    \label{eq:likelihood}
    \text{Pr}(e_i| x_i, \mathfrak{R}(t_i), \beta, \theta )  &=  \sum_{z\in\mathcal{Z}} \frac{e^{{\beta_{z_i}}^{\top} x_i}}{\sum \limits_ {x_j \in {\mathfrak{R}(x_i)}}{} e^{\beta_{z_j}^{\top} x_j } } p(z |  \mathfrak{R}(x_i),  \theta) \\
    &= \mathop{\mathbb{E}}\limits_ {z\sim p(.|\theta)} {} \bigg[ \frac{e^{{\beta_{z_i}}^{\top} x_i}}{\sum \limits_ {x_j \in {\mathfrak{R}(x_i)}}{} e^{\beta_{z_j}^{\top} x_j } }  \bigg]
\end{align}
\begin{align}
    \label{eq:llikelihood}
    \text{and the log-likelihood, }\ell\ell(\theta, \beta; x_i) = \log \Bigg( \sum_{z\in\mathcal{Z}} \frac{e^{{\beta_{z_i}}^{\top} x_i}}{\sum \limits_ {x_j \in {\mathfrak{R}(x_i)}}{} e^{\beta_{z_j}^{\top} x_j } } p(z | \mathfrak{R}(x_i), \theta ) \Bigg)
\end{align}
Where, $ z = \{z_1,z_2, ..., z_{|\mathfrak{R}(x_{i})|} \}$ is the joint set of latent variables corresponding to each of the individuals in the Risk Set, $\mathfrak{R}(x_i)$ 
In order to make the above expectation tractable, \cite{rosen1999mixtures} propose replacing this with, 
\begin{align}
    \widehat{\text{Pr}}(e_i| x_i, \mathfrak{R}(t_i), \beta ) 
    &=    \frac{  \mathop{\mathbb{E}}\limits_ {z_i\sim p(.|\theta, x_i)} {} \big[  e^{{\beta_{z_i}}^{\top} x_i  } \big]}{\sum \limits_ {x_j \in {\mathfrak{R}(x_i)}}{} \mathop{\mathbb{E}}\limits_ {z_j\sim p(z_j|\theta, x_j)} {} \big[  e^{\beta_{z_j}^{\top} x_j  } \big] } \label{eq:roslikelihood}
\end{align}

The denominator in this case requires only $|\mathfrak{R}(x_i)|\times K$ operations to compute. 

\noindent Instead of using the approximation above, we propose to instead use the Variation Lower Bound  of the log-likelihood in \ref{eq:llikelihood}. Notice that the ELBO is tractable, and involves only $|\mathfrak{R}(x_i)|\times K$ operations to compute.
\begin{align}
\label{eq:ELBO}
    \text{ELBO}(\theta, \beta; x_i) &= \mathop{\mathbb{E}}_{ {z_i\sim p(.|\theta, x_i) }} \big[{{\beta_{z_i}}^{\top} x_i} \big] - \log \big( \sum \limits_ {x_j \in {\mathfrak{R}(x_i)}}{}  \mathop{\mathbb{E}}_{ z_j\sim p(.|\theta,  x_j)}    \big[ { e^{\beta_{z_j}^{\top} x_j } } \big] \big)
\end{align}

\noindent A formal proof of this quantity being a lower bound is in the following  Section \ref{sec:theory}

\section{Analysis}
\label{sec:theory}
In this section we provide some theoretical insights for the proposed objective.

\begin{proposition}
\label{prop:factor}
 The objective proposed by \cite{rosen1999mixtures} differs from the likelihood by a factor $\Phi$,\vspace{-0.3em}
 $$\Phi = \bigg[ 1- \frac{1}{\mathbf{B}}\bigg(1 -\bigg[\frac{\mathbf{D}}{\mathbf{B}} - \frac{\mathbf{C}}{\mathbf{A}}  \bigg]\frac{}{} \bigg) \bigg]$$\vspace{-0.3em}
 $$ \quad \mathbf{A} = \mathop{\mathbb{E}}\limits_{z_i \sim p(.|\theta, x_i)}{} [e^{\beta_{z_i}^\top x_i }], \quad \mathbf{B} = \sum _ {x_j \in \mathfrak{R}(x_i)}\mathop{\mathbb{E}}\limits_{z_j \sim p(.|\theta, x_j)}{} [e^{\beta_{z_j}^\top x_j }]$$\vspace{-0.3em}
  $$ \quad \mathbf{C} = \mathop{\mathbb{V}}\limits_{z_i \sim p(.|\theta, x_i)}{} [e^{\beta_{z_i}^\top x_i }], \quad \mathbf{D} = \sum _ {x_j \in \mathfrak{R}(x_i)}\mathop{\mathbb{V}}\limits_{z_j \sim p(.|\theta, x_j)}{} [e^{\beta_{z_j}^\top x_j }]$$ \end{proposition}

\noindent \textit{Proof.} Sketch: The proof involves Taylor's approximation of the fraction around the expectation of the denominator, followed by the independence of each $z_i \in z$.
\begin{align}
\mathbb{E}\bigg[ \frac{X}{Y} \bigg] &\approx \frac{\mathbb{E}[X]}{\mathbb{E}[Y]} \bigg(1 - \frac{\text{Cov}(X, Y)}{\mathbb{E}(X)\mathbb{E}(Y)} + \frac{\mathbb{V}(Y)}{\mathbb{E}(Y)^2} \bigg)\\
\text{Pr}(e_i| x_i, \mathfrak{R}(t_i), \beta ) &=\widehat{\text{Pr}}(e_i| x_i, \mathfrak{R}(t_i), \beta ) \times \Phi
\end{align}

\begin{proposition}
\label{prop:ELBO}
 The $\textup{\textbf{ELBO}}$ in Eq. \ref{eq:ELBO} is a lower bound on the Log-Likelihood in \ref{eq:llikelihood}.
\end{proposition}
\noindent \textit{Proof.} 
\begin{align}
\ell\ell(\theta, \beta; x_i) = \log \bigg( \mathop{\mathbb{E}}\limits_ {z\sim p(.| \mathfrak{R}(x_i), \theta)} {} \Bigg[ \frac{e^{{\beta_{z_i}}^{\top} x_i}}{\sum \limits_ {x_j \in {\mathfrak{R}(x_i)}}{} e^{\beta_{z_j}^{\top} x_j } }  \Bigg] \bigg)
\end{align}
\indent Applying Jensen's Inequality, 
\begin{align}
\ell\ell(\theta, \beta; x_i) &\geq \widetilde{\text{ELBO}}(\theta, \beta; x_i),\label{eq:ineq}\\
\widetilde{\text{ELBO}}(\theta, \beta; x_i) &= \mathop{\mathbb{E}}_{z_i \sim p(.| x_i, \theta) } \big[ \log \big( e^{{\beta_{z_i}}^{\top} x_i} \big) \big] - \mathop{\mathbb{E}}_{z\sim p(.| \mathfrak{R}(x_i), \theta )} \big[ \log \big( {\sum \limits_ {x_j \in {\mathfrak{R}(x_i)}}{} e^{\beta_{z_j}^{\top} x_j } } \big) \big]
\end{align}
\indent This quantity is still intractable to compute since it involves expectation over the joint $z$. Thus, re-applying Jensen's to the second term in reverse, 

\begin{align}
 \widetilde{\text{ELBO}}(\theta, \beta; x_i) &\geq \text{ELBO}(\theta, \beta; x_i), \label{eq:ineq2}\\
\text{ELBO}(\theta, \beta; x_i) &=  \mathop{\mathbb{E}}_{ z_i\sim p({.|x_i, \theta) }} \big[{{\beta_{z_i}}^{\top} x_i} \big] - \log \big( \sum \limits_ {x_j \in {\mathfrak{R}(x_i)}}{}  \mathop{\mathbb{E}}_{ z_j\sim p(.|\mathfrak{R}(x_i),\theta)}    \big[ { e^{\beta_{z_j}^{\top} x_j } } \big] \big)
\end{align}

From Eq. \ref{eq:ineq} and Eq. \ref{eq:ineq2}, $\ell\ell(\theta, \beta; x_i) \geq \text{ELBO}(\theta, \beta; x_i) \qquad \qquad \qquad \qquad \qquad \qquad \qquad \blacksquare $

\begin{proposition}
\label{prop:ELBO1999}
 The $\textup{\textbf{ELBO}}$ in Eq. \ref{eq:ELBO} is also a lower bound on the original objective in \ref{eq:roslikelihood} proposed by \cite{rosen1999mixtures} .
\end{proposition}

\noindent \textit{Proof.} The proof of the above proposition follows immediately from applying Jensen's Inequality in reverse to the first term in Eq. \ref{eq:ELBO}. It can also be derived by using the fact that Model Likelihood and the Objective differ by a factor as derived in Proposition \ref{prop:factor}. $\quad \quad \quad \blacksquare$\\

\noindent The fact that the \textbf{ELBO} is a lower bound on the model likelihood under the DAG assumptions and the original approximation of the likelihood is an interesting result. 

\section{Experiments and Implementation Details} 

We now describe the experimental details of the parameter learning and model inference at test time.

\subsection{Parameter Learning}
The mixture of experts model has two sets of parameters: $\theta$ (for the gating function) and $\beta$ (for the experts). In its simplest form, the gating function is simply a softmax on the inputs and is thus linear, while non-linearities can be introduced using a multi-layered Perceptron. We test the results of our model with both linear and non-linear gating functions. We perform hyperparameter selection via grid-search over multiple activations like \textsf{ReLU} \citep{nair2010rectified}, \textsf{SELU} \citep{klambauer2017self}, Sigmoid, optimization algorithms like \textsf{ADAM} \citep{kingma2014adam} and \textsf{RMSprop} \citep{hinton2012neural}, learning rates, epochs and MLP architectures. We do not include any intercept or bias terms for both the gating as well as the expert set of parameters. For the gating, we hypothesize that not including an intercept term is equivalent to having a uniform prior over the  posterior distribution over the gates conditioned on the input. For the experts, we do not include the intercept as this is common practice with proportional hazard models. We also apply $L_2$ regularization over the parameters of the experts.  We compute the gradients over the entire training dataset, and use them to update the parameters of our model\footnote{Batching data in the presence of risk sets of variable size is challenging and an open problem for further research.}. During the training process, a validation set (held aside from the training data) was leveraged in order to prevent over-fitting the parameters of the network.


\subsection{Inference}

In this section we describe how we deploy the model at test time for data of an unseen individual in order to infer the corresponding hazard value $\lambda(x_i)$ for an individual $x_i$.\\

\noindent \textbf{Soft-Gating:} We compute the expected hazard value under the distribution imposed by the gating function, $f(.)$: 
$$\lambda(x_i) =  \mathop{\mathbb{E}}_{z_i\sim(.|x_i, \theta)} [e^{\beta_{z_i}^\top x_i}] = \sum_k f_k(x_i, \theta) (e^{\beta_k^\top x_i})$$

\noindent \textbf{Hard-Gating:} This involves selecting the expert that has the highest probability under the distribution imposed by the softmax:
$$\hat{k} = \text{arg max}_k f_k(x_i, \theta),\quad \lambda(x_i) =  (e^{\beta_{\hat{k}}^\top x_i})$$

\noindent We would like to point out that hard gating in the presence of a linear softmax function $f(.)$ can be interpretable since the weights on the softmax retain the semantic meaning of the original feature space.

\section{Evaluation Strategy} 
The presence of censoring within survival data makes traditional evaluation metrics like RMSE and R-squared values unsuitable for measuring the performance of survival models. Therefore, a specialized metric, concordance-index, is frequently used to perform model evaluation in survival analysis tasks \cite{DBLP:journals/corr/abs-1708-04649}. Concordance-index can be defined as the ratio of concordant pairs to the total number of comparable pairs within the dataset.
\begin{equation}
c=Pr(\hat{y_i}>\hat{y_j}|y_i>y_j)
\end{equation}
\noindent In the preceding definition, a pair is comparable if either both data points within the pair are uncensored or the observed event time of the uncensored instance is smaller than the censoring time of the censored instance.  
In practice however, concordance index can be calculated in different ways. When the output of the model is a hazard ratio (which is true in our case), concordance-index is computed as follows:
\begin{equation}
\hat{c}=\frac{1}{\text{num}}\sum_{i:e_i=1}\sum_{j:y_i<y_j} \mathbbm{1}[\lambda(x_i, \beta)>\lambda(x_j,\beta)]
\end{equation}
where num is the total number of comparable pairs, i and j $\epsilon$ [1,2...,N], $\mathbbm{1}[.]$ is the indicator function and $\beta$ are the parameters learned for each covariate of the hazard model.

\noindent To report the final results of our models predictions, we bootstrap our test data as described in \cite{efron1994introduction} and calculate the concordance index for each of the 250 bootstrap samples thus generating confidence bands over our method's results.

\section{Datasets} 
We chose three real-world datasets to evaluate the performance of our proposed approach against well-established alternative approaches: CPH from \cite{cox1972regression}, Random Survival Forests (RSF) from \cite{ishwaran2008random} and DeepSurv from \cite{katzman2018deepsurv}.
\subsection{METABRIC (Molecular Taxonomy of Breast Cancer International Consortium)
}
METABRIC from \cite{curtis2012genomic} is a study that aims to classify breast tumors using molecular signatures in order to find the optimal treatment strategy for patients. These molecular signatures are shown to have correlations with the survival of patients. The METABRIC dataset contains clinical information of 1,980 patients and gene expression data. 57.72\% of the patients observe death due to breast cancer over the duration of the study. We consider covariates including age at diagnosis, hormone treatment indicator, radiotherapy indicator, chemotherapy indicator and ER-positive indicator with 4 gene indicators: (MKI67, EGFR, PGR, and ERBB2) to evaluate our model.
 
\subsection{SUPPORT (Study to Understand Prognoses Preferences Outcomes and Risks of Treatment)}
The SUPPORT dataset from \cite{knaus1995support} was created to study the survival times of seriously ill and hospitalized adults. A prognostic model to estimate survival over a 180-day period was developed as a result of this study. The SUPPORT dataset set contains information on 14 covariates  (age, sex, race, number of comorbidities, presence of diabetes, presence of dementia, presence of cancer, mean arterial blood pressure, heart rate, respiration rate, temperature, white blood cell count, serum’s sodium, and serum’s creatinine) of 9,105 patients. 68\% of the patients under observation observe death during the study. The cohort had a median death time of 58 days. 
\subsection{Rotterdam-GBSG (Rotterdam and German Breast Cancer Study Group)}
The Rotterdam tumor bank was leveraged to evaluate the prognostic importance of four major components of urokinase-type plasminogen activator (uPA) system of plasminogen in the prognosis of several types of cancers, \cite{foekens2000urokinase}. We use the information of 1,546 patients with node-positive breast cancer to train our models. 90\% of these patients observe death during this study. We evaluate and compare our models by leveraging testing our performance on the GBSG dataset from \cite{schumacher1994randomized} that contains information on 686 patients. 56 percent of the data is censored. Data preprocessing follows the guidelines provided by \cite{altman2000we}

\section{Results} 

This section compares the efficacy of our proposed variational lower bound to the approximation of the objective \textbf(RT-MoCE) as proposed by \cite{rosen1999mixtures}.  We also compare the results that our method achieves with hard-gating and soft-gating in comparison to competing baselines. We observe that we beat the reported concordance for all the datasets using the proposed approach.

\subsection{Comparison to RT-MoCE}

\begin{figure}[!htbp]
    \centering
    \begin{minipage}{0.33\textwidth}
    \centering
    \textsc{Rotterdam-Gbsg}
    \includegraphics[width=\textwidth]{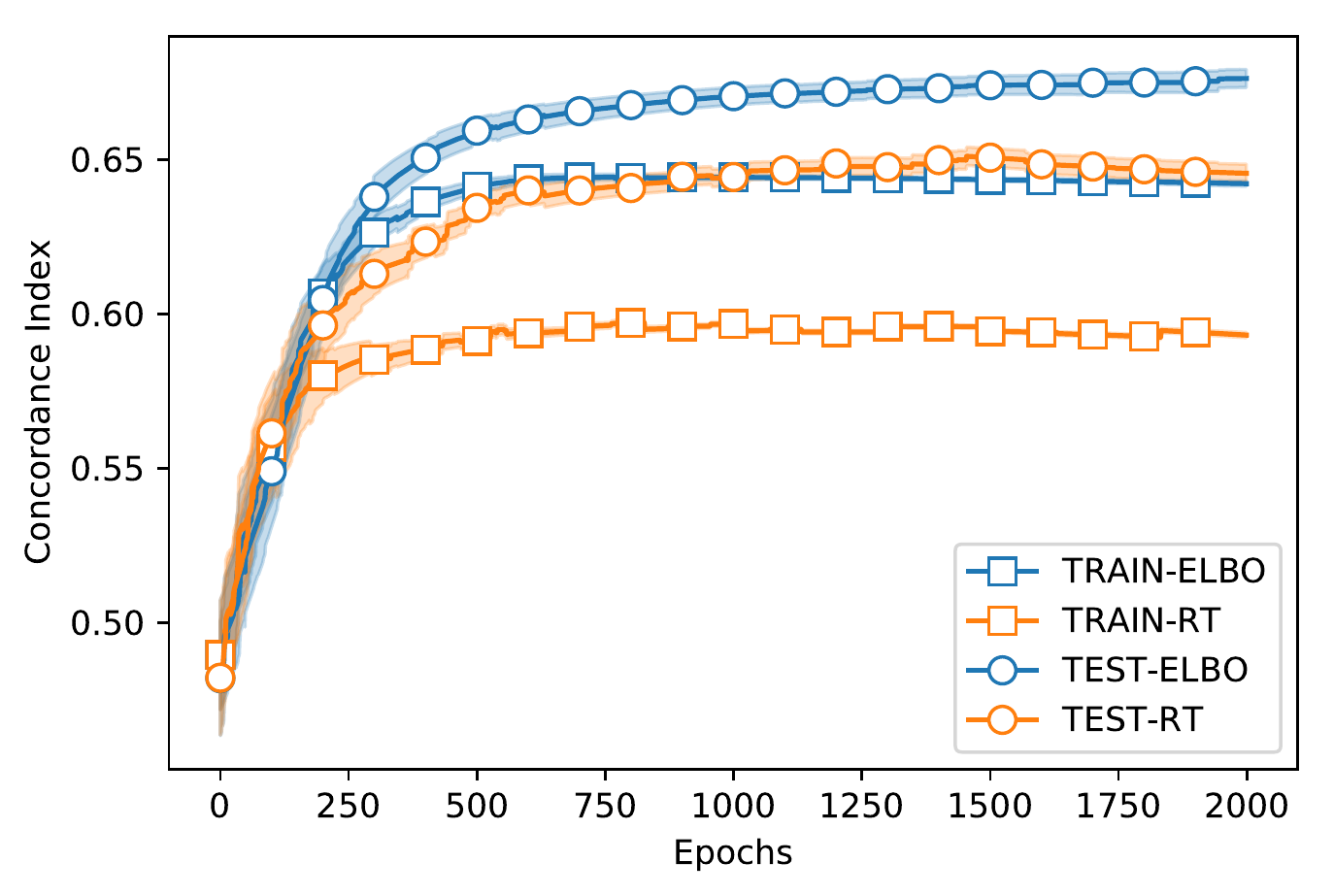}
    \end{minipage}%
    \begin{minipage}{0.33\textwidth}
    \centering
    \textsc{Metabric}
    \includegraphics[width=\textwidth]{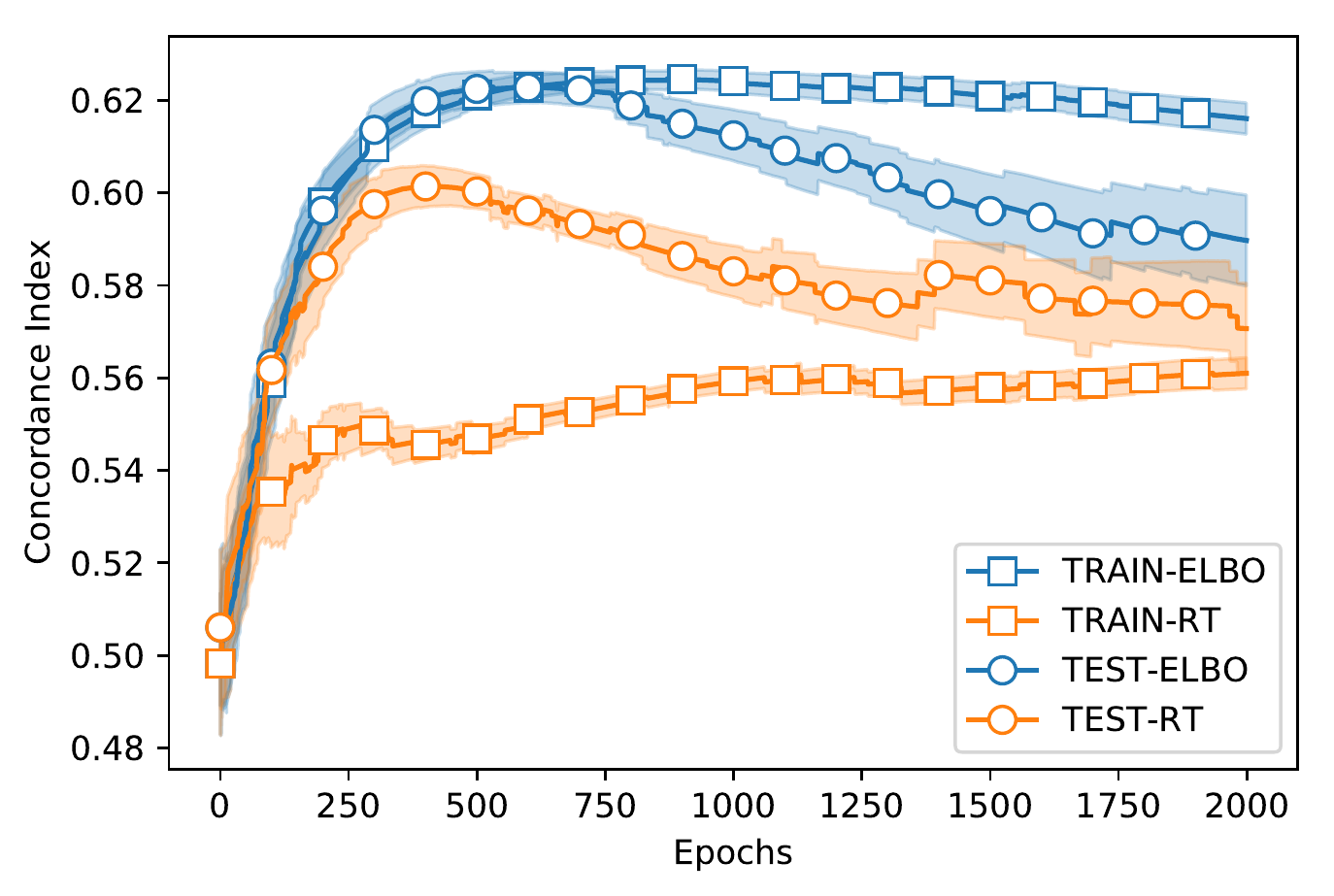}
    \end{minipage}
    \begin{minipage}{0.33\textwidth}
    \centering
    \textsc{Support}
    \includegraphics[width=\textwidth]{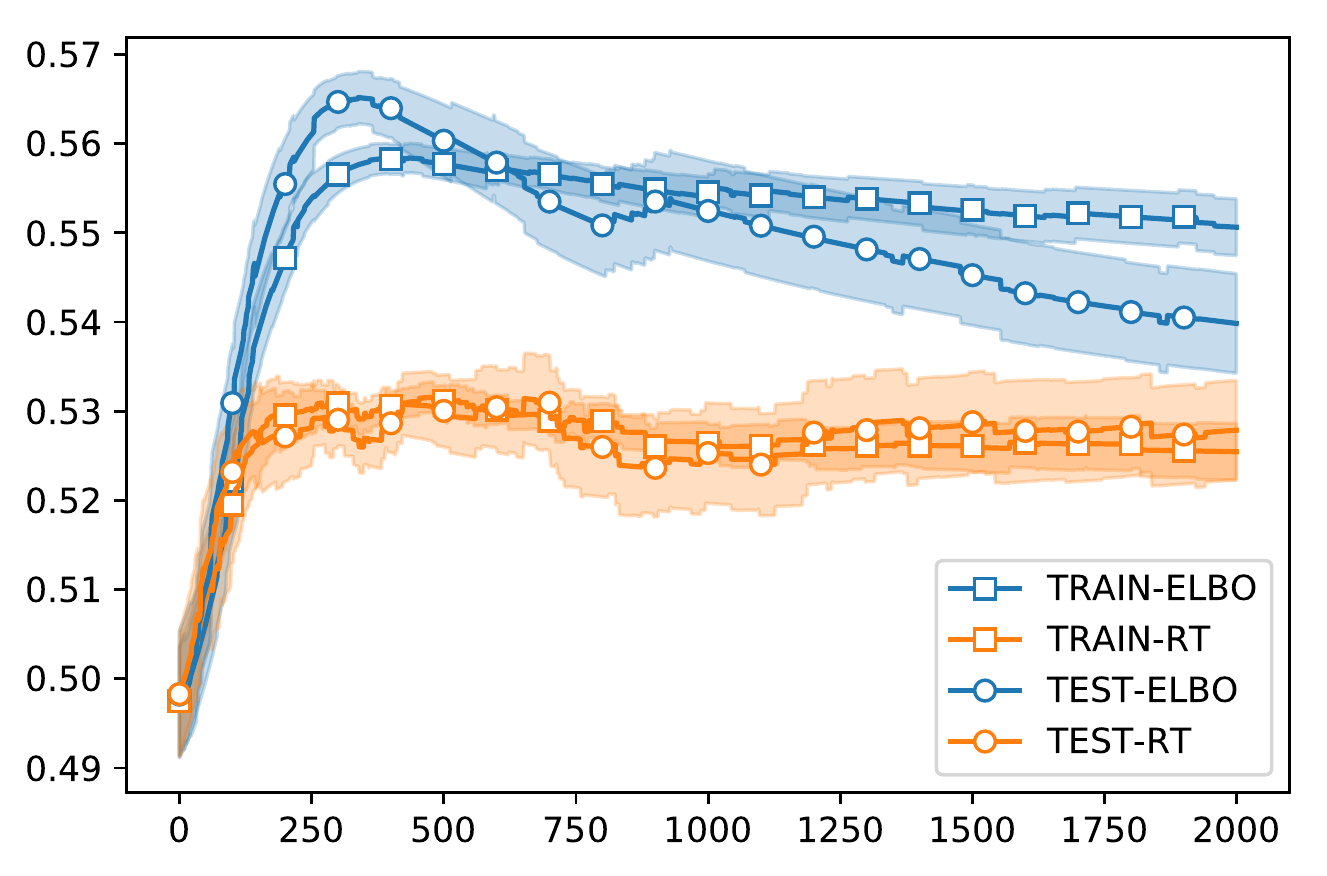}
    \end{minipage}
    \caption{Concordance-index using Hard-Gating for proposed objective in comparison to the RT-approximation over train and test. (Error Bands repsent the 95\% CI)}
    \label{fig:rtmoe}
\end{figure}

The Figure \ref{fig:rtmoe} compares the concordance index  at each epoch of training our Cox expert mixture using the proposed variational lower bound versus the RT-MoCE. Note that the we use hard gating for inference to compute the Concordance Index.
 For each of the datasets we perform 100 different random initializations and train the model for 2000 epochs and average across each run. 
 Notice also that across all the three datasets, our proposed loss outperforms the RT-MoCE loss connsiderably which demonstrates that out approach is robust to overfitting.

\subsection{Performance against Baselines}

\begin{table}[!htbp]
  \centering 
 
\begin{tabular}{|c|c|c|c|}\toprule \midrule
\textsc{ Model} &  \textsc{Metabric} &\textsc{ Rotterdam-Gbsg} &\textsc{ Support} \\ \hline
\textsc{Cph} & $0.6306\pm0.004$ & $0.6578\pm0.004$ & $0.5828\pm0.002$\\ 
\textsc{Rsf} & $0.6243\pm0.004$ & $0.6512\pm0.003$ & $0.6130\pm0.002$\\ 
\textsc{DeepSurv} & $0.6434\pm0.004$ & 0.$6684\pm0.003$ & $0.6183\pm0.002$ \\ 
\rowcolor{Gray} \textsc{Elbo-MoCE-HG} & $0.6349\pm0.003$ & $\mathbf{0.6866\pm0.002}$ & $0.5706\pm0.001$ \\ 
\rowcolor{Gray}\textsc{Elbo-MoCE-SG} & $\mathbf{0.6585\pm0.003}$ & $0.6752\pm0.002$ & $\mathbf{0.6196\pm0.001}$ \\ \midrule\bottomrule
  \end{tabular}
   \caption{Results of hard and soft linear gating networks and their comparison with relevant baselines (95\% bootstrap CI ).}
     \label{tab:results1}
\end{table}

\begin{table}[!htbp]
  \centering 
\begin{tabular}{|c|c|c|c|}\toprule\midrule
\textsc{ Model} &  \textsc{Metabric} &\textsc{ Rotterdam-gbsg} &\textsc{ Support} \\ \hline
\rowcolor{Gray}\textsc{Elbo-MoCE-HG} & $0.6373\pm0.003$ & $\mathbf{0.6825\pm0.002}$ & $0.5775\pm0.001$ \\ 
\rowcolor{Gray}\textsc{Elbo-MoCE-SG} & $\mathbf{0.6658\pm0.003}$ & $0.6815\pm0.002$ & $\mathbf{0.6203\pm0.001}$ \\ \midrule \bottomrule
  \end{tabular}
   \caption{Results of hard and soft nonlinear gating networks and their comparison with relevant baselines (95\% bootstrap CI).} 
     \label{tab:results2}
\end{table}

\noindent We compare the output of the MoCE architecture with our proposed ELBO Loss (\textsc{Elbo-MCoE}) against the parametric baselines like plain Cox regression (\textsc{CPH}) and \textsc{DeepSurv} as well as the non-parametric Random Survival Forest (\textsc{RSF}) in terms of Concordance Index (Tab.~\ref{tab:results1} and \ref{tab:results2}). Our experimental protocol is to have an 80/20 train/test split. We also holdout a small subset of the training data for validation to assess overfitting. We report the Concordance Index on the held out test set for both \textsc{MoCE} with linear and deep nonlinear gating functions. 

\noindent From Tab.~\ref{tab:results1}, it is clear that our approach substantially outperforms existing state of the art models, and retains interpretability in terms of a linear gating function. 
At the cost of a little loss of interpretability\footnote{Notice we say a `little loss' of interpretability since even though the gating function is nonlinear, the final expert that makes the decision is a Linear function that is interpretable in the semantics of the original feature representation.} with the nonlinear gating function, we can further push the performance numbers in terms of Concordance Index, as is apparent from results in Tab.~\ref{tab:results2}.



\noindent Code repository available at: \url{https://github.com/rohansangave/SurvivalAnalysis}

\section{Discussion and Conclusion}
We have revisited the technique of easing the linearity constraint of Cox regression for survival analysis by adding nonlinearities to the model using a mixture of experts. We have done so by leveraging a variational lower bound of the log-likelihood under the Mixture of Cox Expert modelling assumptions, without which parameter learning would be intractable. As a result, we have developed a semi-parametric model that while being interpretable has competitive performance against relevant baselines that includes current state of art deep learning techniques. 

 We validated our claims through several experiments on real-world clinical datasets. Through this paper we aim to bring to light the experimental evidence that suggests deep learning may not be ideal solution for problems involving structured prediction where interpretability is paramount. 
 
 We further conjecture that the assumption of a constant baseline hazard across the entire population is a strong one, and that we can improve our performance in survival prediction by easing this assumption. In the future we propose to experiment with models that make parametric assumptions on the baseline hazards themselves and hypothesize that conditioning the baselines hazards on an individual's covariates would allow us to push the performance of semi-parametric survival even further.

\bibliographystyle{abbrv}
\bibliography{ref}
\bigskip

\appendix

\section{Hyperparameters for the trained models.}

\begin{table}[htbp]
  \centering 
  \caption{Hyperparameters for different datasets with hard-gating network} 
  \begin{tabular}{|l|l|l|l|}\hline
  {\bfseries Hyper-parameter} & 
   {\bfseries Metabric} &
    {\bfseries GBSG} &
    {\bfseries SUPPORT} \\ \hline
    Optimizer & Adam & Adam & Adam \\ \hline
    Learning rate & 0.001 & 0.001 &  0.001\\ \hline
    \# epochs & 4000 & 4000 & 4000 \\ \hline
    \# linear models & 10 & 12 & 10\\ \hline
    \# hidden layer & 1 & 0 & 2\\ \hline
    Hidden layer sizes & [ 9 ] & [  ] & [ 14, 14 ] \\ \hline
  \end{tabular}
  \label{tab:example} 
\end{table}

\begin{table}[htbp]
  \centering 
  \caption{Hyperparameters for different datasets with soft-gating network} 
  \begin{tabular}{|l|l|l|l|}\hline
  {\bfseries Hyper-parameter} & 
   {\bfseries Metabric} &
    {\bfseries GBSG} &
    {\bfseries SUPPORT} \\ \hline
    Optimizer & Adam & Adam & Adam \\ \hline
    Learning rate & 0.0001 & 0.001 & 0.001\\ \hline
    \# epochs & 4000 & 4000 & 4000\\ \hline
    \# linear models & 12 & 5 & 5\\ \hline
    \# hidden layer & 2 & 1 & 1 \\ \hline
    Hidden layer sizes & [ 9, 9 ] & [ 7 ] & [ 14 ]\\ \hline
  \end{tabular}
  \label{tab:example} 
\end{table}

\end{document}